\documentclass[sigconf,nonacm]{acmart}

\usepackage{booktabs}
\usepackage{graphicx}
\usepackage{tikz}
\usetikzlibrary{positioning, arrows.meta, fit}

\title{Behavioral Fingerprints for LLM Endpoint Stability and Identity}

\author{Jonah Leshin}
\author{Manish Shah}
\author{Ian Timmis}
\affiliation{
  \institution{Project VAIL}
  \city{San Francisco}
  \country{USA}
}
\email{jonah@projectvail.org}
\author{Daniel Kang}
\affiliation{
  \institution{University of Illinois Urbana-Champaign}
  \department{Department of Computer Science}
  \city{Urbana}
  \state{Illinois}
  \country{USA}
}

\begin{abstract}
The consistency of AI-native applications depends on the \emph{behavioral
consistency} of the model endpoints that power them. Traditional reliability
metrics such as uptime, latency and throughput do not capture
behavioral change, and an endpoint can remain
``healthy'' while its effective model identity changes
due to updates to weights, tokenizers, quantization,
inference engines, kernels, caching, routing, or
hardware. We introduce \textbf{Stability Monitor}, a
black-box stability monitoring system that periodically
fingerprints an endpoint by sampling outputs from a fixed
prompt set and comparing the resulting output
distributions over time. Fingerprints are compared using
a summed energy distance statistic across
prompts, with permutation-test p-values as evidence of distribution shift
aggregated sequentially to detect
change events and define stability periods. In controlled
validation, Stability Monitor detects changes to
model family, version, inference stack, quantization, and behavioral parameters. In
real-world monitoring of the same model hosted by
multiple providers, we observe substantial
provider-to-provider and within-provider stability
differences.
\end{abstract}


\begin{document}
\maketitle

\section{Introduction}

\subsection{Reliability is not stability}
The software deployment intuition ``if it's up and
fast, it's fine'' is insufficient for AI-native systems: an
endpoint can pass standard health checks while its outputs
drift, shifting formatting, tone, or tool behavior in
ways that break downstream parsers and guardrails.
Stability failures are particularly impactful in multi-step agentic
workflows, where small variance in early steps can
compound into large workflow variance. This
motivates \textbf{stability} as a distinct operational
metric: behavioral consistency of a model endpoint
under a fixed interface contract.

Even if users fix visible settings (e.g., temperature),
endpoint behavior can vary due to system level nondeterminism ``beneath the
surface'': variance in inference engines, kernels, caching, batch sizes, and hardware---and providers may
route requests across heterogeneous environments. The values of these factors
at run time may be a function of overall system load, which the user cannot control,
thus making the endpoint nondeterministic from the user's
perspective \cite{he2025nondeterminism}. 

These system-level factors, which can result in
nondeterministic behavior even at temperature 0, must be considered on top of
discrete, deliberate updates such as a model version
change, undermining reproducibility in both production
and evaluation environments. Evaluation instability and provider effects
are now recognized as a major source of variance in
benchmarking results \cite{epoch2025benchmarking}.

\subsection{Contribution and scope}
We introduce two components: \textbf{Stability Monitor}, which
continuously fingerprints endpoints and detects behavioral change
events, and \textbf{Stability Arena}, a web application that
publishes and visualizes the data Stability Monitor produces.
Together, the goal is to operationalize continuous stability
monitoring under realistic constraints:
\begin{itemize}
  \item \textbf{Black-box}: observe only public API
  I/O; no privileged access to weights or provider
  internals.
  \item \textbf{Lightweight and frequent}: monitor at
  high cadence versus heavy periodic evals (motivating the notions of
  stability periods and change events).
  \item \textbf{Actionable}: produce audit trails and
  change event alerts usable by engineering,
  security, and compliance.
\end{itemize}

Stability Arena exposes the system's
monitoring capabilities---built on its
fingerprinting methodology and sequential
change-detection pipeline---through a live web interface
that lets users inspect these capabilities against
real endpoint data.

We contribute:
\begin{enumerate}
  \item A practical fingerprinting and change-detection
  pipeline based on energy-distance permutation testing
  and sequential evidence aggregation.
  \item Controlled validation across multiple real-world
  change types.
  \item Real-world provider comparisons for the same
  nominal model showing large cross-provider and
  within-provider stability differences, motivating
  ``same model, different behavior'' as an operational
  reality.
\end{enumerate}

\section{Background and related work}

Prior work on model fingerprinting includes frameworks for intellectual property
protection-- verifying whether a deployed model is a copy or derivative
of a protected model (\cite{cao2021ipguard,uchida2017embedding}). These methods typically require white-box
access or adversarial inputs crafted to distinguish specific
architectures. Our work differs in both goal and setting: rather than
verifying model ownership, we detect \emph{behavioral change} at an
endpoint over time, using only standard black-box API access with
natural-language prompts.

Chen et al.\ \cite{chen2023chatgpt} demonstrated that the behavior of
GPT-3.5 and GPT-4 changed substantially over several months, with
measurable shifts in accuracy, formatting, and safety behavior,
providing direct motivation for continuous behavioral monitoring of
production endpoints.

More recent work on black-box change detection includes B3IT
\cite{chauvin2026b3it}, which uses per-endpoint ``border
inputs''---prompts where multiple tokens are nearly tied as the top
prediction at low temperature. Like our approach,
B3IT frames the problem as a hypothesis test on the output distribution
and targets continuous monitoring. A key difference is that B3IT's
probes are tailored to each endpoint during an initialization phase
and may require re-discovery after a change event, since the
model's decision boundaries may have shifted. Our system uses
a fixed, model-agnostic prompt set that persists across change events
and endpoints, avoiding per-endpoint initialization.

We use an energy distance statistic as the core metric
comparing samples of response embeddings. This metric is a
nonparametric distance between distributions and is
zero if and only if the distributions match under mild conditions (as
a metric on distributions of random vectors)
\cite{szekely2013energy}. Energy distance is closely related to
maximum mean discrepancy (MMD) \cite{gretton2012kernel}; we chose
energy distance for its interpretability and parameter-free
formulation.

We use sequential evidence accumulation based on
e-values (expectation-bounded test statistics) to
enable continuous monitoring with optional stopping
behavior, suited for streaming detection settings
\cite{wang2023tinyreview,ramdas2025evalues}.

Our statistical approach to measuring behavioral change is 
complementary to projects that run domain-specific evaluations over time
\cite{marginlab2025tracker,phillips2025dailybench}.
These efforts offer a signal about specific abilities over time,
but are typically expensive to run, limiting the feasibility of coverage
across a range of models and providers.

\section{Problem definition}

In the context of text-to-text models, we define
\textbf{endpoint stability} as distributional
consistency of responses under a fixed prompt set and
fixed user-visible settings. We treat the endpoint as a black box.
Endpoint instability may occur due to changes to:
\begin{itemize}
  \item model identity
  (family, version, tokenizer, weights),
  \item deployment (inference engine, stack),
  \item optimization techniques (quantization, caching,
  kernels, speculative decoding),
  \item configurations (sampling parameters,
  system prompts)
\end{itemize}

Our objective is to detect when there has been a change in the endpoint's
effective output distribution and to create an audit trail of
\textbf{change events} and \textbf{stability periods}.

\section{Methodology: fingerprints and change
detection}

\subsection{Fingerprint construction}
A fingerprint is constructed by selecting a fixed set
of prompts and sampling multiple responses per prompt;
responses are embedded as real-valued vectors. A
fingerprint thus consists of a \emph{set of sets of
vectors}---one sample set per prompt. Our implementation requires a total
of 800 inference requests, each consisting of a few tokens, to be made to an endpoint
to generate a fingerprint.

\subsection{Pairwise fingerprint comparisons}
Given fingerprints $X$ and $Y$, we compute an energy
distance statistic per prompt between the
prompt-specific embedding samples, and sum
across prompts to produce an aggregate distance $E = E(X, Y)$.

To test the null hypothesis that two fingerprints are drawn from the
same underlying distribution, we compute a p-value via a permutation test on the
aggregate statistic $E$. To compute the p-value, the embedding
vectors for each prompt are pooled and randomly re-split into two
groups of the original sizes according to
a permutation $\pi$ on the pooled set of vectors, yielding a permuted statistic
$E_\pi$. The p-value is the fraction of permutations for which
$E_\pi \geq E$ when this comparison is done across a set of randomly selected
permutations. Lower p-values indicate stronger evidence of change.

\subsection{Fingerprint comparisons over time}
Monitoring of an endpoint begins by generating a baseline fingerprint $F_0$. 
At a regular cadence we generate new fingerprints $F_i$ at time $t_i$ and 
compute the aforementioned p-value $p_i = p(F_0, F_i)$. Each time a new
fingerprint $F_i$ is generated, using the e-values methods from \cite{wang2023tinyreview,ramdas2025evalues} applied
to the sequence of p-values $p_1, ..., p_i$,
we assess the cumulative evidence for a change in the
endpoint's effective output distribution since the time of the baseline fingerprint.
If the evidence exceeds a predetermined threshold, we declare a
change event and set the most recent
fingerprint as the new baseline, to which subsequent fingerprints are compared.

\begin{figure*}[t]
  \centering
  \begin{tikzpicture}[
    box/.style={draw, thick, rounded corners=2pt, minimum height=1cm,
                align=center, inner sep=6pt},
    group/.style={draw, dashed, rounded corners=4pt, inner sep=10pt},
    >=Stealth
  ]

  \node[box] (pa) {Provider A};
  \node[box, below=5pt of pa] (pb) {Provider B};
  \node[box, below=5pt of pb] (pc) {Provider C};
  \node[group, fit=(pa)(pb)(pc),
        label={above:\textbf{Inference Providers}}] (pgroup) {};

  \node[box, right=2.8cm of pb, minimum height=2.2cm, minimum width=3.6cm,
        align=center] (svc)
       {\textbf{Stability Monitor}\\[3pt]
        \small generate $F_i$, compare $F_i$ vs $F_0$\\
        \small detect change events};
  \node[anchor=north] at (svc.south) {\small(VM)};

  \node[box, right=2.5cm of svc, align=left, inner sep=8pt] (db)
       {\textbf{Database}\\[3pt]
        \small $\bullet$~fingerprints\\
        \small $\bullet$~p-values\\
        \small $\bullet$~change events};

  \node[box, right=2.2cm of db] (arena) {\textbf{Stability}\\{\textbf{Arena}}};
  \node[anchor=north] at (arena.south) {\small arena.projectvail.com};

  \draw[->, thick] ([yshift=3pt]svc.west) -- ([yshift=3pt]pgroup.east)
    node[above, midway] {request};
  \draw[->, thick] ([yshift=-3pt]pgroup.east) -- ([yshift=-3pt]svc.west)
    node[below, midway] {response};

  \draw[->, thick] (svc.east) -- (db.west)
    node[above, midway] {write};

  \draw[->, thick] (db.east) -- (arena.west)
    node[above, midway] {read};

  \end{tikzpicture}
  \caption{System architecture. Stability Monitor periodically queries
  inference providers, generates fingerprints $F_i$, compares them
  against the baseline $F_0$ to compute p-values and detect change
  events, and writes all results to the database. The Stability Arena
  web application reads from the database to display monitoring data.}
  \label{fig:pipeline}
\end{figure*}
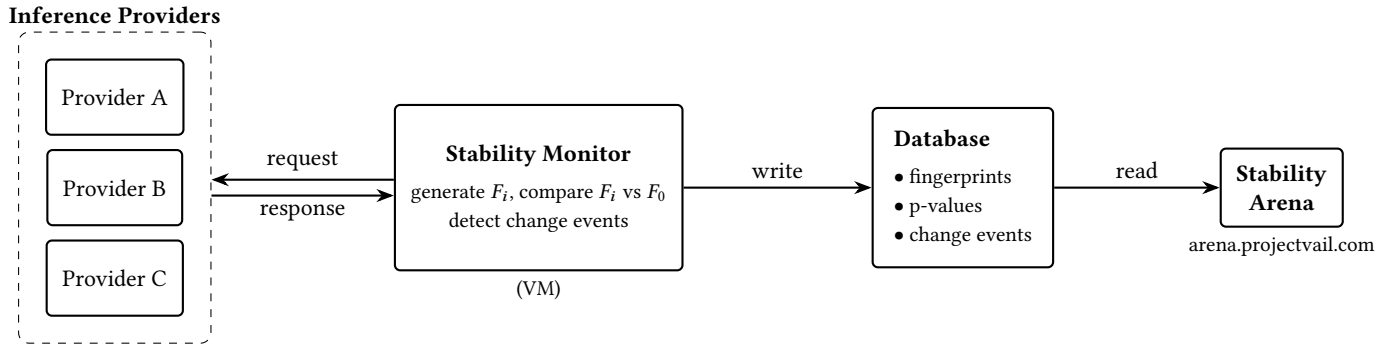

\section{System: Stability Monitor and Stability Arena}
Stability Monitor runs continuously against an
endpoint:
\begin{enumerate}
  \item generate baseline fingerprint $F_0$,
  \item at regular cadence (typically every few hours) generate $F_i$,
  \item compute $p_i = p(F_i, F_0)$,
  \item update sequential evidence and detect change
  events,
  \item write an audit log of stability periods and
  events.
\end{enumerate}

Stability Arena (\url{https://arena.projectvail.com}) publishes
stability data across endpoints and supports
cross-provider comparisons for the same nominal model. It runs as
a service that makes API requests to model providers to perform inference generation.
Endpoint metadata, fingerprints, p-values, and change events are stored in a cloud
postgres database. Stability Arena runs as a separate service that pulls
data from the database and displays it in a web interface.

The web interface exposes several views. A provider matrix
shows cumulative change event counts per endpoint over
time, giving an at-a-glance picture of which providers
have been most and least stable. Per-endpoint views
surface the full change event history alongside
traditional operational metrics such as latency and
error rates, situating behavioral stability in the
context of conventional reliability signals. A
cross-provider comparison view shows each provider's
deviation from the pooled behavior of all providers
serving the same nominal model, making it straightforward
to identify outlier provider behavior. Because Stability
Arena runs continuously against live endpoints, the data
reflects the latest real behavior rather than a static snapshot.

\section{Findings: controlled methodology validation}

\subsection{Experimental setup}
Validation experiments were run by hosting a model at a
local endpoint and generating fingerprints via API calls. In an independent
process, we ran Stability Monitor, generating fingerprints at an hourly cadence.
Once Stability Monitor had generated a few fingerprints, we made a discrete
change to the model hosted at the local endpoint without modifying Stability
Monitor. Stability Monitor continued to run, generating fingerprints,
unaware of the change that had been made to the endpoint.

We tested five change types, with examples of each shown in Table \ref{tab:validation}.
Each change was evaluated on its own, holding the other change types constant.

\begin{table}[t]
  \centering
  \caption{Validation tests.}
  \label{tab:validation}
  \begin{tabular}{ll}
    \toprule
    Change type & Example Intervention \\
    \midrule
    Model family & Qwen $\to$ Llama \\
    Version upgrade & Qwen2.5-0.5B $\to$ Qwen3-0.6B \\
    Inference stack & vLLM $\to$ Transformers \\
    Quantization & Qwen2.5-14B BF16 $\to$ Qwen2.5-14B INT8 \\
    Temperature & 0.7 $\to$ 0.6 \\
    \bottomrule
  \end{tabular}
\end{table}

\subsection{Results}
With one exception, each change produced a change event
immediately on the next fingerprint; the exception was
a small temperature change
(0.7 $\rightarrow$ 0.6), which took 18 fingerprints after the change was made to
trigger a change event.

Moreover, Stability Monitor recorded a single change
event per intervention: we observed stability before the change and
stability after detection relative to the new baseline fingerprint.

\section{Findings: cross-provider comparisons for the
same model}
VAIL's fingerprints support comparisons of different endpoints
serving the same nominal model in two modes:
(a) pairwise comparisons over shared time windows, and
(b) individual endpoint divergence relative to a group of endpoints.

\subsection{Pairwise provider similarity (heatmap)}
\begin{figure}[t]
  \centering
  \includegraphics[width=\linewidth]{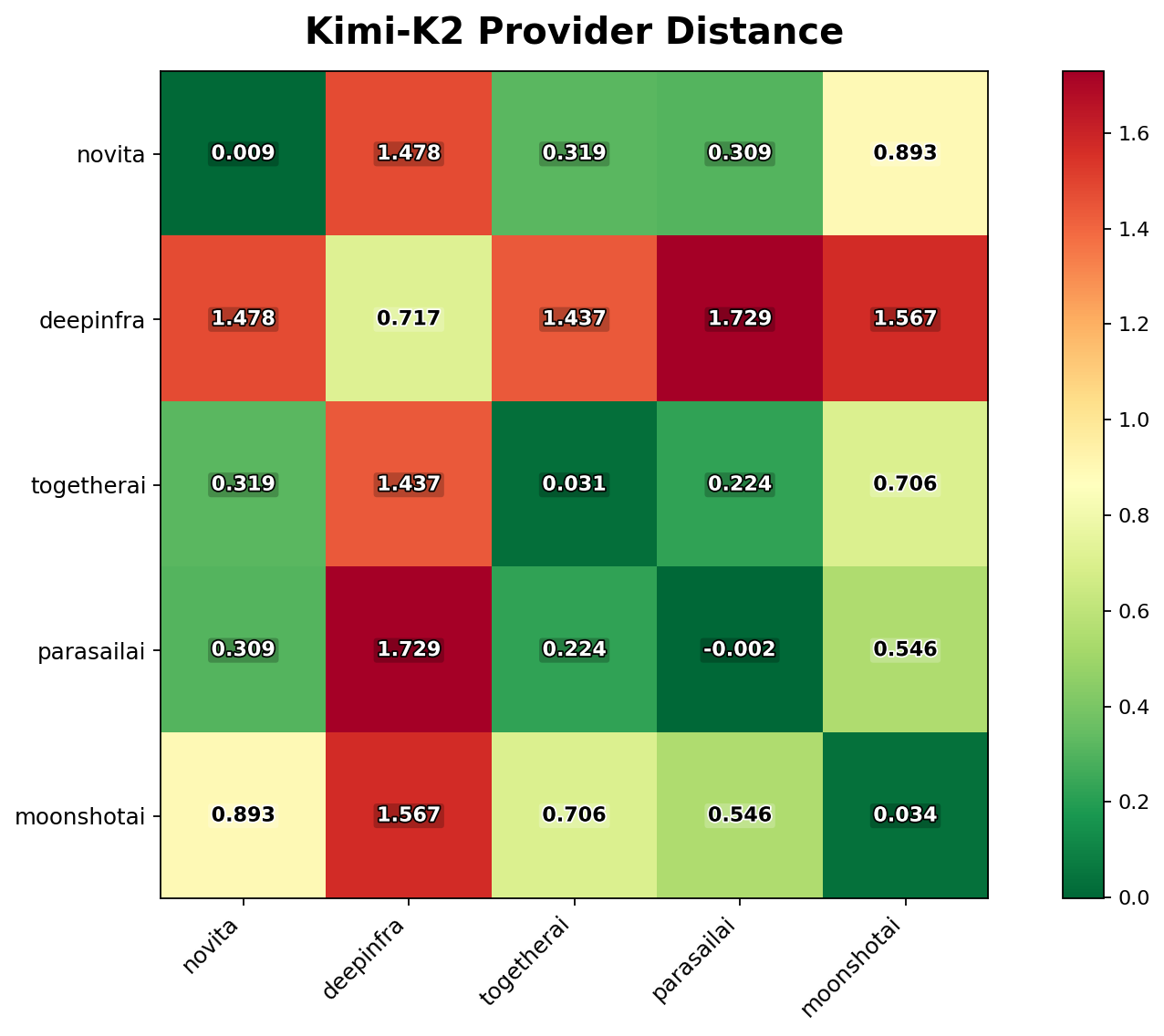}
  \caption{Comparison of selected providers serving Kimi-K2-0905-Instruct from
  2025-11-27 to 2025-11-28.}
  \label{fig:providersim}
\end{figure}

In the case of (a), we consider multiple providers serving the same nominal model
over the same time period. During this time period, we perform pairwise fingerprint
comparisons between each pair of providers using energy distance to form
a distance matrix. 
One application is provider identification: given an endpoint serving a model
known to be among a set of candidate providers, we can determine which provider
is serving it by fingerprinting the endpoint and comparing against the candidates.
In Figure \ref{fig:providersim}, the diagonal entries are the minimum in each row and column,
confirming that a provider's fingerprints are most similar to its own---and therefore
that nearest-neighbor comparison against known-provider fingerprints reliably
identifies the serving provider.\footnote{A diagonal entry can be
slightly negative because the finite-sample unbiased energy statistic
$2\,\mathbb{E}\|X{-}Y\| - \mathbb{E}\|X{-}X'\| -
\mathbb{E}\|Y{-}Y'\|$ can dip below zero when sampling noise causes
the within-fingerprint spread terms to slightly exceed the
cross-fingerprint term; in the population limit the statistic is
non-negative and equals zero when the distributions match.}
We can likewise perform the above analysis over a sequence of consecutive
shared time windows, enabling us to measure pairwise provider similarity over time.

\subsection{Individual endpoint divergence}
For the case of (b), we propose a ``provider vs the pack''
deviation score in which we compare one provider's fingerprints to
the pooled distribution of all other providers'
fingerprints for the same model over matched time windows. For each provider,
we compute the distance between that provider and an aggregate distribution of all other providers' fingerprints.
We then normalize these distances relative to the median distance across all providers to the aggregate distribution
to produce a divergence ratio. For example, a provider with a divergence ratio of 2 is twice as different
from the aggregate distribution as the median provider is.

In the Stability Arena "Provider Comparisons" tab we compute this
divergence ratio daily and plot its values over
time, enabling us to identify which providers diverge from the consensus and
when. Because the metric is normalized by the median, it is sensitive to
individual outliers rather than uniform shifts; a rolling model update that
propagates unevenly across providers would appear as a sequence of outlier spikes.

\section{Findings: real-world deployment stability examples}
In production monitoring of the Kimi-K2-0905-Instruct
model across providers, Stability Monitor observed that
in November 2025, one provider (DeepInfra) was least stable---nearly every
fingerprint generation resulted in an output distribution 
that triggered a change event. On the other hand, the endpoint hosted by the
model's creator (Moonshot) showed 100\% stability.

In December 2025, when Stability Monitor flagged a model change event for
Parasail, the Parasail team confirmed a
hardware-provider switch due to a physical node
failure.

\section{Caveats and limitations}

Even for a fixed model, inference execution specifications
(engine, kernels, caching, speculative decoding,
hardware) can affect next-token probabilities and token
selection; providers may route/batch requests across
heterogeneous environments, inducing ongoing randomness
that complicates attribution and can blur the line
between ``change event'' and ``always somewhat
unstable.''

This presence of ongoing, infrastructure-driven randomness is consistent with systems work arguing that
endpoint nondeterminism is often driven by
load-induced batch-size variation interacting with
non-batch-invariant kernels
\cite{he2025nondeterminism}. Stability Monitor is
designed to detect endpoint-level changes, and
endpoint-level variance can be a product of these
execution-level factors.

\section{Conclusion}
Behavioral instability raises security and compliance
concerns: if a model changes silently, prior safety
validation and guardrails may no longer apply. Stability monitoring provides
an audit trail of when behavior changed.

Operationally, Stability Arena also surfaces that
``the same model'' served by different providers can
exhibit different behavioral characteristics due to
provider infrastructure choices; this matters for
selecting production providers and for interpreting
benchmark or eval results that depend on provider
access paths.

\bibliographystyle{ACM-Reference-Format}
\bibliography{stability_acm_refs}

\end{document}